\documentclass[final,5p,times,twocolumn]{elsarticle}

\usepackage{amssymb}
\usepackage{amsmath}
\usepackage{graphicx}
\usepackage{multirow}
\usepackage{longtable}
\usepackage{rotating}
\usepackage{color}
\usepackage{makecell}
\usepackage[colorlinks,
bookmarks=false,
linkcolor=black,
anchorcolor=black,
citecolor=black
]{hyperref}
\usepackage{cite}
\usepackage{bm}
\usepackage{lineno}
\graphicspath{{figures/}}

\journal{XXXX}

\begin{document}

\begin{frontmatter}

\title{CrossFuse: A Novel Cross Attention Mechanism based Infrared and Visible Image Fusion Approach}


\author[a]{Hui Li\corref{correspondingauthor}}

\author[a]{Xiao-Jun Wu}


\address[a]{International Joint Laboratory on Artificial Intelligence of Jiangsu Province, \\ School of Artificial Intelligence and Computer Science, \\ Jiangnan University, 214122, Wuxi, China \fnref{address1}}

\cortext[correspondingauthor]{Corresponding author email: lihui.cv@jiangnan.edu.cn}

\begin{abstract}
Multimodal visual information fusion aims to integrate the multi-sensor data into a single image which contains more complementary information and less redundant features. However the complementary information is hard to extract, especially for infrared and visible images which contain big similarity gap between these two modalities. The common cross attention modules only consider the correlation, on the contrary, image fusion tasks need focus on complementarity (uncorrelation). Hence, in this paper, a novel cross attention mechanism (CAM) is proposed to enhance the complementary information. Furthermore, a two-stage training strategy based fusion scheme is presented to generate the fused images. For the first stage, two auto-encoder networks with same architecture are trained for each modality. Then, with the fixed encoders, the CAM and a decoder are trained in the second stage. With the trained CAM, features extracted from two modalities are integrated into one fused feature in which the complementary information is enhanced and the redundant features are reduced. Finally, the fused image can be generated by the trained decoder. The experimental results illustrate that our proposed fusion method obtains the SOTA fusion performance compared with the existing fusion networks. The codes of our fusion method will be available soon.

\end{abstract}



\end{frontmatter}


\section{Introduction}

The aim of image fusion is to improve the visual quality of images and provide more accurate and reliable information for various applications \citep{liu2020multi}\citep{zhang2021deep}. multimodal image fusion plays a very important role in computer vision filed and industrial area \citep{zhang2021image}\citep{vivone2023multispectral}, which involves combining information from different imaging modalities such as visible, infrared etc. Specially, with the development of vision sensors, how to utilize the benefits of these multimodality data to sever the real world scenario becomes a crucial problem. To this end, a lot of researchers work on the visual information fusion task and many milestone achievements have been made, such as multi-scale transformer \citep{pajares2004wavelet} \citep{li2013image}, sparse representation \citep{liu2016image}\citep{li2017multi}, pre-trained deep learning models \citep{li2018infrared}\citep{liu2017multi}, Bayesian based fusion model \citep{zhao2020bayesian}. multimodal image fusion also has numerous applications in other fields such as medical diagnosis \citep{tang2022matr}\citep{zhou2023gan}, surveillance \citep{voronin2022deep}\citep{yadav2023contrast}, remote sensing \citep{wang2022review}\citep{ma2023multimodal}, and robotics \citep{liang2022deep}\citep{liu2023mff} etc.

Deep learning, as a widely popular technology, is no exception in multimodal image fusion field \citep{li2018densefuse}\citep{ma2019fusiongan}\citep{ma2022swinfusion}\citep{Zhao_2023_CVPR}\citep{li2023lrrnet}. Thanks to the multimodal datasets \citep{hwang2015multispectral}\citep{liu2022target}, the deep learning based fusion methods have emerged as a promising approach due to their ability to learn complex feature representations from source images and effectively fuse multiple images. These methods can be broadly classified into two categories: multi-stage fusion and end-to-end fusion. 

\begin{figure}[!ht]
    \centering
    \includegraphics[width=\linewidth]{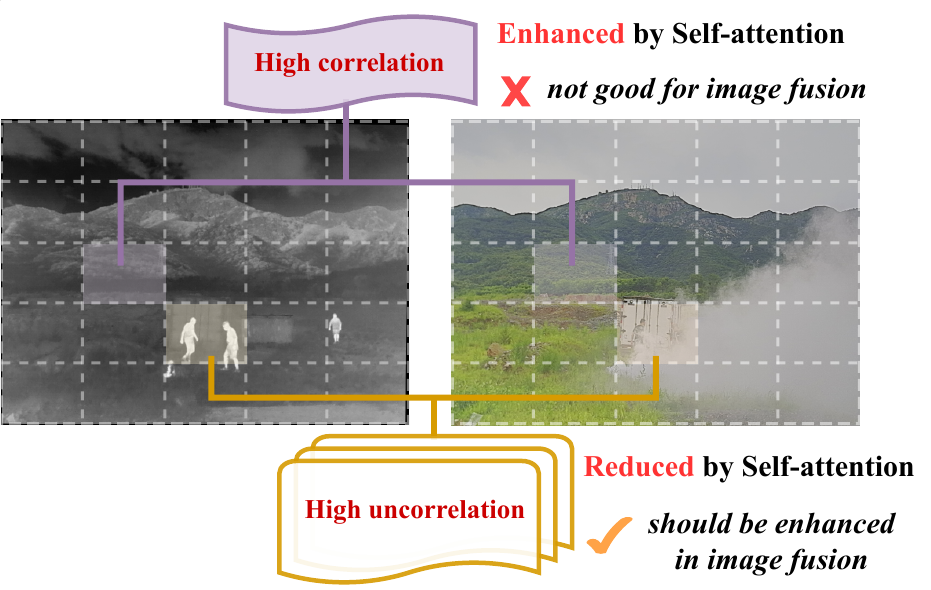}
    \caption{The correlation and uncorrelation for self-attention in image fusion task. For multimodal images, the self-attention may not be suitable for inter-modality  processing. In image fusion task, the redundant information will be enhanced and the complementary features are reduced, which is more obvious when the source images are all in gray-scale.}
    \label{fig:first}
\end{figure}

Multi-stage fusion methods \citep{li2018densefuse}\citep{zhao2020DBLP}\citep{li2021rfn}\citep{wang2022swinfuse} involve separate processing of the input images before combining them, whereas end-to-end fusion methods \citep{xu2020u2fusion}\citep{li2023lrrnet}combine the input images directly. Both types of methods have their strengths and disadvantages. Multi-stage fusion methods tend to be more flexible and adaptable, allowing for the incorporation of a wide range of pre-processing techniques. However, they are often computationally intensive and can suffer from information loss during the processing stages. 

Multi-stage image fusion methods divide the fusion process into several stages \citep{li2018densefuse}\citep{li2021rfn}. Each stage handles a particular aspect of the fusion process, such as feature extraction, feature fusion and image generation. In this kind of approaches, deep learning models can be injected in each stage, such as convolutional neural networks (CNNs) for feature extraction \citep{liu2016image}\citep{li2018infrared}\citep{li2019infrared}, a specific light network is trained to replace the handcrafted fusion strategy \citep{li2021rfn}. These methods tend to be more flexible and adaptable, allowing for the incorporation of a wide range of pre-processing techniques. However, it requires careful selection of the models, fusion strategies and training strategies to ensure optimal results.

End-to-end fusion methods \citep{xu2020u2fusion}\citep{zhang2020ifcnn}, on the other hand, are more efficient and can preserve more information without the manual operations. These methods involve designing a single deep learning model that takes multiple input images and directly generates fused image. The fusion networks are utilized to extract features and learn how to fuse them in a single step \citep{zhao2022efficient}\citep{tang2022ydtr}. One of the advantages of these approaches is that the optimal fusion strategy can be learned from the input images without requiring any prior knowledge of the fusion process. However, it can be more challenging to design an effective end-to-end fusion model, and it requires a carefully designed loss function to achieve optimal performance \citep{wang2023interactively}. Unfortunately, these disadvantages are also the crucial problems in most computer vision tasks.

To address the above disadvantages, transformer architecture, as a key technology, has been introduced into multimodal image fusion task \citep{ma2022swinfusion}\citep{vs2022image}\citep{wang2022swinfuse}\citep{zhang2023transformer}. Transformer-based methods have shown promising performance in various tasks, including natural language processing \citep{2017Attention} and image classification \citep{Dosovitskiy2021An}\citep{liu2021swin}. These approaches use self-attention mechanism to capture global dependencies and facilitate efficient feature representation learning. Current transformer-based fusion methods also follow this scheme. However, these methods primarily concentrate on the self-attention mechanism and ignore the interplay between various modalities \citep{tang2022ydtr}\citep{wang2022swinfuse} \citep{qu2022transmef} \citep{tang2023datfuse}. However, the complementary information between different modalities is the key for the multimodal fusion task, thus cross-attention should be paid more attention.

Correlation among samples is a fundamental aspect of computer vision fields, as it reflects the significant features for various tasks \citep{afyouni2022multi}\citep{zhu2022visual}\citep{li2022multi}. However, in image fusion fields, particularly multimodal image fusion, complementary (uncorrelation) information is crucial \citep{zhang2021image}\citep{vivone2023multispectral}. Therefore, the uncorrelation should be paid more attention in image fusion tasks. Current transformer-based fusion methods \citep{tang2023datfuse}\citep{zhang2023transformer}\citep{TANG2023tccfusion} focus solely on the self-attention mechanism, which is the primary component of transformer. As shown in Fig.\ref{fig:first}, while this mechanism can improve the correlation between inputs, it may also reduce the complementary information. 

In recent studies, while the self-attention mechanism has demonstrated the capability to enhance complementary information through well-designed loss functions and feature fusion modules, it is important to note that mishandling feature correlation can result in a degradation of fusion performance in specific scenarios.


To overcome the limitations of current transformer-based fusion techniques, in this paper, a novel cross-attention mechanism (CAM) based fusion method is proposed, which employs self-attention to enhance the intra-features of each modality while utilizes cross-attention based architecture to enhance the inter-features (complementary information) between different modalities. By injecting the CAM into the transformer architecture, the proposed method offers a powerful fusion strategy for multimodal images that effectively enhances the uncorrelation between inputs. The main contributions of this paper are summarized as follows,

1. A cross-attention mechanism is introduced to enhance multimodal features in this paper. The proposed mechanism optimizes the fusion process by effectively augmenting complementary features, resulting in outcomes that are both more accurate and comprehensive.

2. A novel hybrid fusion network is presented in this research, amalgamating the strengths of convolutional layers with attention mechanisms (both self and cross) for the multimodal image fusion task. This methodology facilitates the extraction of deep features from source images, maintaining detail information, and enhancing complementary information.

3. In comparison to state-of-the-art fusion methods, the experimental results demonstrate that the method proposed in this paper presents a promising alternative to current fusion techniques. It furnishes a more robust and efficient solution for the multimodal image fusion task.

The rest of our paper is structured as follows. In Section \ref{sec-relate}, we briefly review the related work on deep learning-based fusion. The proposed fusion framework is described in detail in Section \ref{sec-proposed}. The experimental results are presented in Section \ref{sec-experiments}. Finally, we draw the paper to conclusion in Section \ref{sec-con}.

\section{Related works}
\label{sec-relate}

In this section, two key techniques based methods are briefly introduced, including: transformer based fusion methods and cross-attention based methods.

\subsection{Transformer based fusion methods}

Transformer is a deep learning architecture originally developed for natural language processing tasks \citep{2017Attention}\citep{floridi2020gpt}, it has also been successfully applied to computer vision fields \citep{Dosovitskiy2021An}\citep{qu2022transmef}\citep{TANG2023tccfusion}. The key innovation is the self-attention mechanism, which allows the model to weigh the importance of different input elements when making predictions.

In computer vision, the transformer can be used in various ways. One common approach is to use it as an alternative to CNNs for feature extraction, it also appears in image fusion task \citep{maaz2023edgenext}\citep{yuan2023effective}. Instead of using a fixed kernel to extract local features, the transformer computes attention weights between all pairs of input elements, allowing it to capture global dependencies and relationships among the features. This approach has also been shown to be effective for image fusion tasks \citep{zhou2022multi}\citep{chen2023shape}.

\begin{figure*}[!ht]
	\centering
	\includegraphics[width=0.85\linewidth]{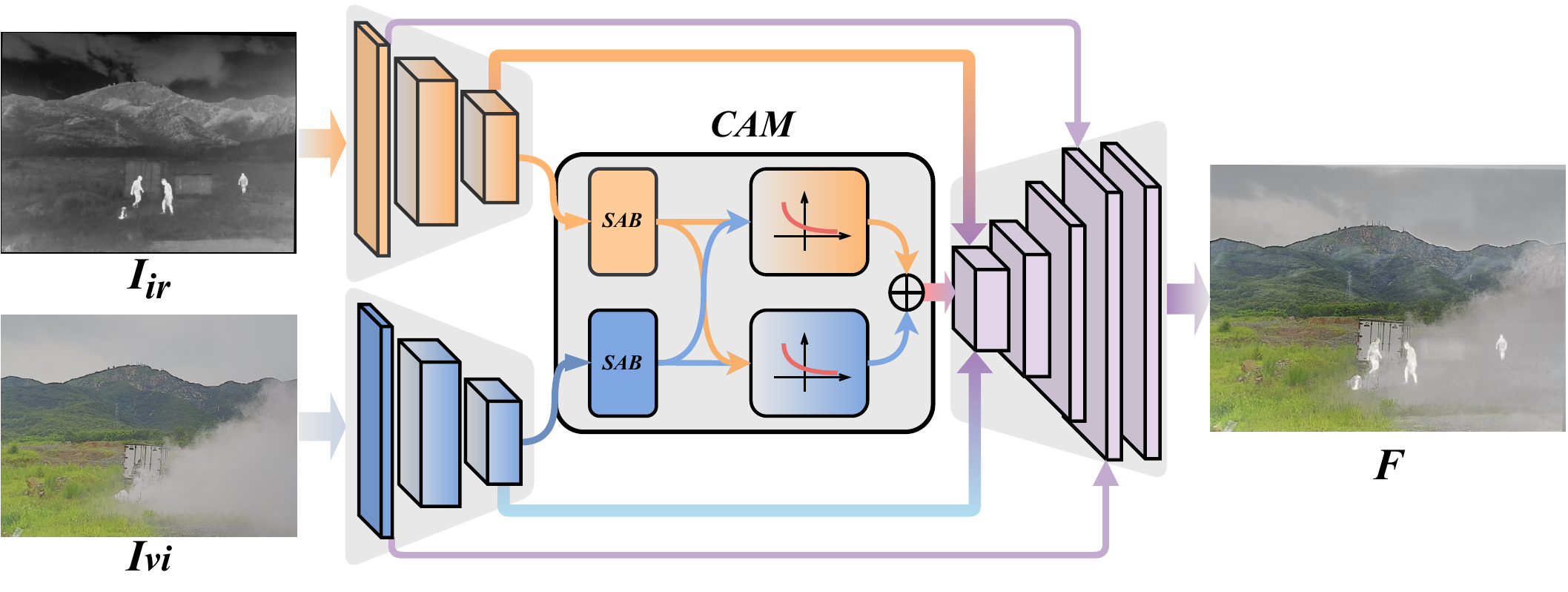}
	\caption{The framework of CrossFuse. Two ``Encoder" contain same architecture but different parameters. The cross-attention mechanism (CAM) is utilized to fuse the multimodal features. ``SAB'' indicates the self-attention block. The fused image can be obtained by ``Decoder" with the long connection from encoders.}
	\label{fig:framework}
\end{figure*}

Although these transformer-based fusion methods obtain good fusion performance, the drawbacks are still observed: (1) The transformer architecture is only utilized in feature extraction stage or image reconstruction stage, the relations between different modalities are not considered \citep{wang2022swinfuse}\citep{qu2022transmef}; (2) Even with the self-attention mechanism in feature fusion stage, these methods still do not address the crucial problem which is that the self-attention mechanism may reduce the complementary information \citep{ma2022swinfusion}\citep{jha2023gaf}.

\subsection{Cross-attention based fusion methods}

Cross-attention is a technique used in computer vision tasks that primarily focuses on the interaction of information between different modalities \citep{ma2019locality}\citep{zhu2022dual}\citep{praveen2022joint}. It is often utilized in multimodal tasks where information from different modalities needs to be integrated to solve a specific problem, such as image fusion \citep{kim2022multi}\citep{zhou2022cafnet}\citep{rao2023gan}, and image registration based fusion method \citep{tang2022superfusion}\citep{xie2023semantics}. 

In transformer architecture, cross-attention is also a key concept, which has been shown to be effective for integrating information from multiple modalities \citep{huang2019ccnet}\citep{zhu2022dual}. In transformer, cross-attention is computed between the features of the encoder and the decoder, where the encoder produces one modality features and the decoder produces another modality features. There are also some fusion methods combine the cross-attention and transformer to obtain better performance \citep{ma2022swinfusion}\citep{jha2023gaf}. However, these cross-attention only focuses on the correlation and ignores the complementary information.

Although the cross-attention mechanism has received widespread attention in image fusion task, the relationship between attention mechanism and the fundamental issue of fusion task has not been fully explored. Thus, how to design an appropriate cross-attention mechanism which preserves the complementary information is crucial.

\section{The proposed method}
\label{sec-proposed}
The proposed CAM based fusion network focuses on the fundamental problem of image fusion task, in which the cross-attention mechanism in image fusion task should enhance the complementary (uncorrelation) information and reduce the redundant (correlation) features. In this section, the architecture and the loss function are introduced in detail.

\subsection{The architecture of the fusion network}
\label{afn}
The architecture is shown in Fig.\ref{fig:framework}. $I_{ir}$ and $I_{vi}$ indicate the infrared image and visible image, respectively. Two encoders are utilized to extract multimodal features from source images. The CAM based transformer architecture is introduced to fuse the multimodal features. Finally, the fused image is generated by decoder. There are two skip connections between encoder and decoder, which are utilized to preserve more deep features and shallow features from source images.

\subsubsection{Encoder architecture}

\begin{figure}[!ht]
	\centering
	\includegraphics[width=0.8\linewidth]{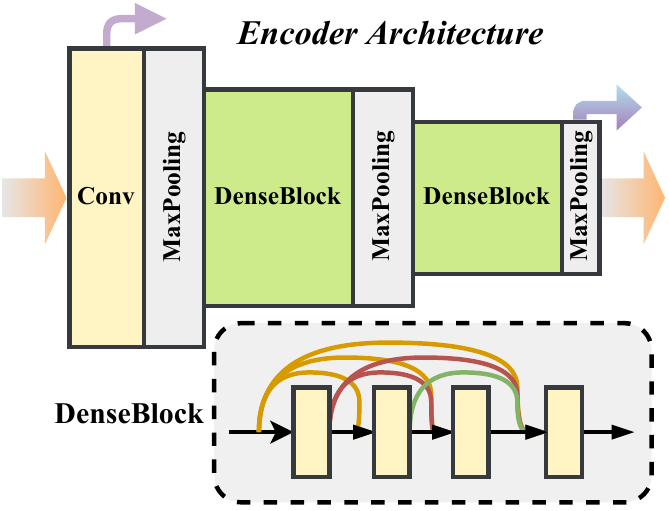}
	\caption{The encoder architecture which contains three blocks: ``Conv", ``MaxPooling" and ``DenseBlock". ``Conv" indicates one convolutional layer, ``DenseBlock" includes four convolutional layers with dense connection.}
	\label{fig:encoder}
\end{figure}

Considering the gap between two modalities (infrared and visible), it is natural to extract the features with different parameters. Thus, in our framework, two encoders with the same architecture but different parameters are utilized. The architecture of encoder is shown in Fig.\ref{fig:encoder}.

The first convolutional layer, ``Conv", is utilized to extract the shallow features from source images, which contains rich texture information. Following the pooling operation, MaxPooling, and the multi-scale features are exploited and the features will preserve more useful information with DenseBlock. With the deeper of encoder, the extracted deep features will focus on salient contents. 

Furthermore, to enhance the detail information and salient features, two skip connections (Conv and the last DenseBlock follow the MaxPooling) are applied into encoder and decoder.

\subsubsection{Cross-attention mechanism (CAM)}
\label{sec:cam}

\begin{figure}[!ht]
	\centering
	\includegraphics[width=\linewidth]{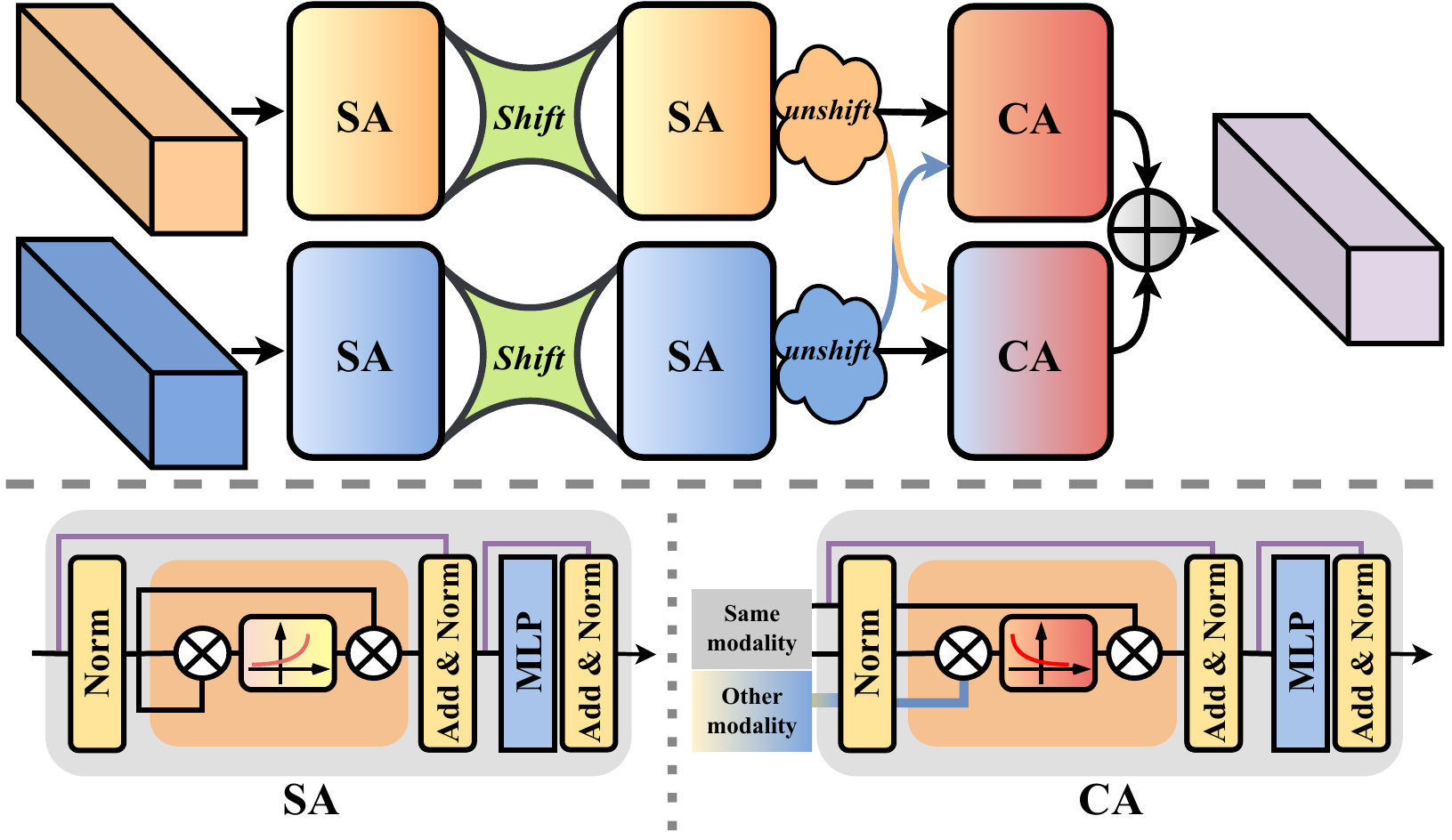}
	\caption{The cross-attention mechanism architecture. ``SA" follows the standard transformer architecture which contains one self-attention block. The ``Shift" and ``unshift" mean the block shift and shift back operation. ``CA" indicates the novel cross-attention mechanism which focuses on the uncorrelation information.}
	\label{fig:cam}
\end{figure}

The proposed cross-attention mechanism (CAM) is introduced in this section, which is the most important part in our method. The architecture of CAM is shown in Fig.\ref{fig:cam}.

Two branches with different parameters are utilized to extract features from two modalities. Each modality features are fed to self-attention (SA) block first to enhance the intra-features, which fits the insight of SA. To further enhance the intra-features, the shift operation is also introduced into CAM, in which the positions of features are moved horizontally and vertically. Then, another SA block is used to enhance the shifted features which will contain more global information. Before the cross-attention, the ``unshift" is utilized to restore the positions. Thus, there are twice as many SA as CA.

After obtaining the intra-enhanced features, the proposed cross-attention block is introduced. The formulas of SA are given as follows,
\begin{eqnarray}\label{equ:sa}
\begin{split}
    &[Q_c, K_c, V_c] = x^c U_{qkv}, \\
    &x_{sa}^c = x_{sa}^c + norm(softmax(\frac{Q_c K_c^T}{\sqrt{d}}) V_c),\\
    &x_{sa}^c = x_{sa}^c + MLP(norm(x_{sa}^c)), \\
    &s.t. \quad U_{qkv}\in \mathbb{R}^{d\times 3d}, c\in{\{ir, vi\}}
\end{split}
\end{eqnarray}
where $x^c$ means the input of SA, $Q_c$, $K_c$ and $V_c$ indicate the different representation of input, $d$ is the dimension of the input vector. $U_{qkv}$ is a transformation matrix which can be learned by a fully connection layer, $norm(\cdot)$ indicates the linear norm operation, $MLP(\cdot)$ means the multilayer perceptron. 

The formulas of CA are given as follows,
\begin{eqnarray}\label{equ:ca}
\begin{split}
    &[Q_{\hat{c}}, K_c, V_c] = [x^{\hat{c}}, x^{c}, x^{c}]\mathcal{U}_{qkv}, \\
    &x_{ca}^{c} = x_{ca}^{c} + norm(re\text{-}softmax(\frac{Q_{\hat{c}} K_c^T}{\sqrt{d}}) V_{c}),\\
    &x_{ca}^{c} = x_{ca}^{c} + MLP(norm(x_{ca}^{c})), \\
    &s.t. \quad \mathcal{U}_{qkv}\in \mathbb{R}^{3d\times 3d}, c\in{\{ir, vi\}}, \hat{c}\neq c
\end{split}
\end{eqnarray}
where the $\hat{c}$ and $c$ indicate the different modality.

The main difference between SA and CA is that the activation function after the matrix multiplication. For different modalities, the complementary (uncorrelation) information rather than the redundant (correlation) features should be enhanced. Thus, a new activation function, reversed softmax ($re\text{-}softmax$), is embed into our cross-attention mechanism, the formulation is given as follows,
\begin{eqnarray}\label{equ:resoftmax}
\begin{split}
    re\text{-}softmax(X)&=softmax(-X)
\end{split}
\end{eqnarray}

\begin{figure}[!ht]
	\centering
	\includegraphics[width=0.6\linewidth]{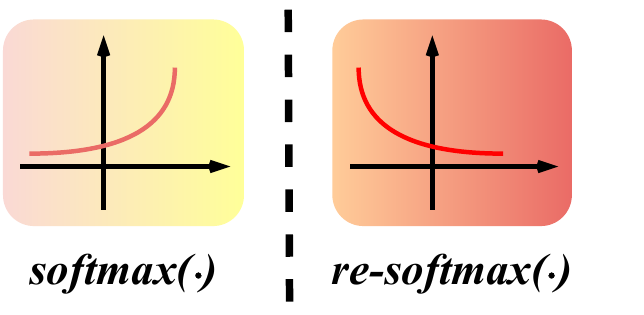}
	\caption{The activation function curves of $softmax(\cdot)$ and $re\text{-}softmax(\cdot)$.}
	\label{fig:curve}
\end{figure}

The activation function curves of $softmax(\cdot)$ and $re\text{--}softmax(\cdot)$ are shown in Fig.\ref{fig:curve} which shows the trend of different activation functions. With the $re\text{--}softmax(\cdot)$, our CA block can focus on the uncorrelation information between different modalities.

\subsubsection{Decoder architecture}

\begin{figure}[!ht]
	\centering
	\includegraphics[width=0.8\linewidth]{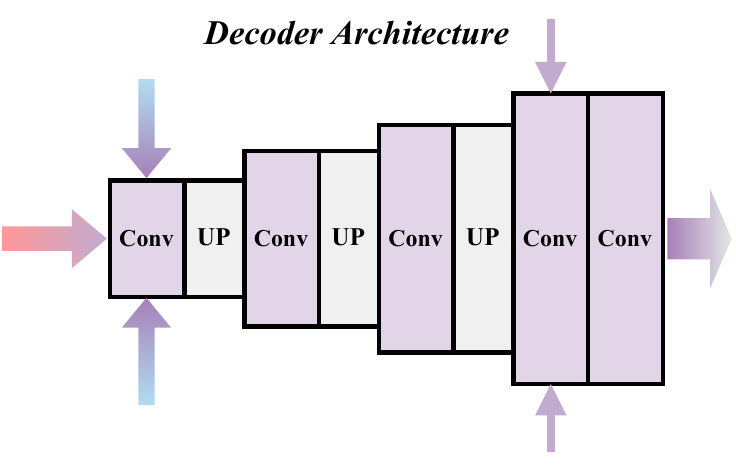}
	\caption{The decoder architecture.}
	\label{fig:decoder_f}
\end{figure}

After CAM, a decoder network is introduced into our framework to obtain the final fused image. In this decoder, several convolutional layers and up-sampling operations are included. The architecture is shown in Fig.\ref{fig:decoder_f}.

To better enhance the salient features and preserve more detail information from source image features, two skip connections between encoder and decoder are introduced into our network, where the deep feature connection for salient features and the shallow connection for detail information. In addition, the feature intensity aware strategy is applied into decoder for multi-level feature fusion, the formula is defined as follows,
\begin{eqnarray}\label{equ:decoder}
\begin{split}
    \Phi_{df}^m&=\Phi_c^m+w_{ir}^m \Phi_{ir}^m+w_{vi}^m \Phi_{vi}^m, \quad m\in{\{deep, shallow\}} \\
    s.t.&\quad w_{ir/vi}^m(\cdot) = \frac{\nabla^m \Phi_{ir/vi}^m(\cdot)}{\sum_{\hat{t}\in{\{ir, vi\}}}\nabla^m \Phi_{\hat{t}}^m(\cdot)}
\end{split}
\end{eqnarray}
where $(\cdot)$ means the position in deep features, $\Phi_c^m$ indicates the features extracted by CAM, $\Phi_{ir}^m$ and $\Phi_{vi}^m$ denote the features from source images (infrared and visible). $\nabla^m$ denotes the detail and base information extractor for shallow features and deep features, respectively. The formulas of $\nabla^m$ are given as follows, 
\begin{eqnarray}\label{equ:detal}
    \begin{split}
        \nabla^{deep}\Phi &= k_\nabla \otimes \Phi \\
        \nabla^{shallow}\Phi &= \sqrt{(1 - k_\nabla \otimes \Phi)^2}
    \end{split}
\end{eqnarray}

\begin{eqnarray}\label{equ:kernel}
k_\nabla = \frac{1}{9}
\left[\begin{array}{c c c}
1 & 1 & 1 \\
1 & 1 & 1 \\
1 & 1 & 1 
\end{array}\right]
\end{eqnarray}
where $\otimes$  $k_\nabla$ is the kernel of convolutional operation to extract the base information. Given the Eq.\ref{equ:detal} and Eq.\ref{equ:kernel}, $\nabla^{deep}$ focuses on the salient features and $\nabla^{shallow}$ enhances the detail information.

\subsection{Training phase with two-stage strategy}

To train our fusion framework, a two-stage training strategy\citep{li2021rfn} is applied. Firstly, an auto-encoder network is constructed for each modality (infrared and visible), which is utilized to reconstruct the inputs. Then, with the trained encoders for each modality, the proposed CAM and decoder are trained with the multimodal data and the proposed loss function.

\subsubsection{First stage for encoders}

\begin{figure}[!ht]
	\centering
	\includegraphics[width=\linewidth]{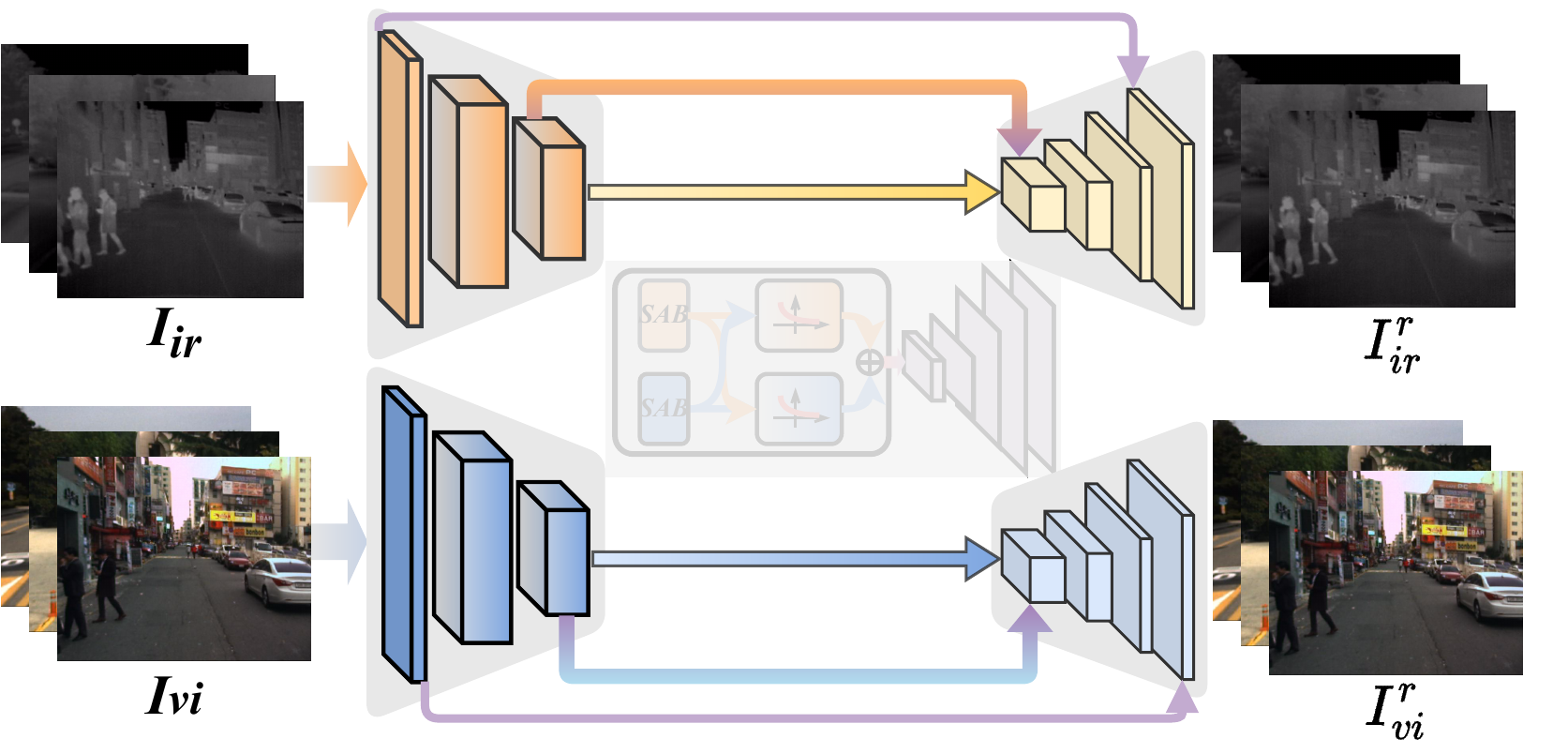}
	\caption{The first training strategy. Two auto-encoders are trained to reconstruct the inputs(infrared images and visible images).}
	\label{fig:train_auto}
\end{figure}

In first stage, the encoders are trained to extract rich features which are benefit for generating the fused image, the framework is shown in Fig.\ref{fig:train_auto}. Since there are feature gap between infrared and visible, it is reasonable to train different parameters for each modality.

As shown in Fig.\ref{fig:train_auto}, these two auto-encoders have same network structure but different parameters. Two skip connections are utilized to preserve the shallow features (detail) and deep features (salient).

Furthermore, to train the auto-encoder network, pixel-level loss ($||\cdot||_F^2$) and structural similarity loss ($SSIM$) are introduced. The loss function for auto-encoders is given as follows,
\begin{eqnarray}\label{equ:loss-auto}
\begin{split}
    L_{auto}^c = ||I_{c}-I_{c}^r||_F^2 + w_s SSIM(I_{c},I_{c}^r), \quad c\in{\{ir,vi\}}
\end{split}
\end{eqnarray}
where $I_{c}^r$ indicates the reconstructed image along with the certain modality(infrared or visible), $w_s$ denotes the trade-off parameter which is set to $1e4$.

Finally, we only use the trained encoders to extract deep features from corresponding modality.

\subsubsection{Second stage for CAM and decoder}

\begin{figure}[!ht]
	\centering
	\includegraphics[width=\linewidth]{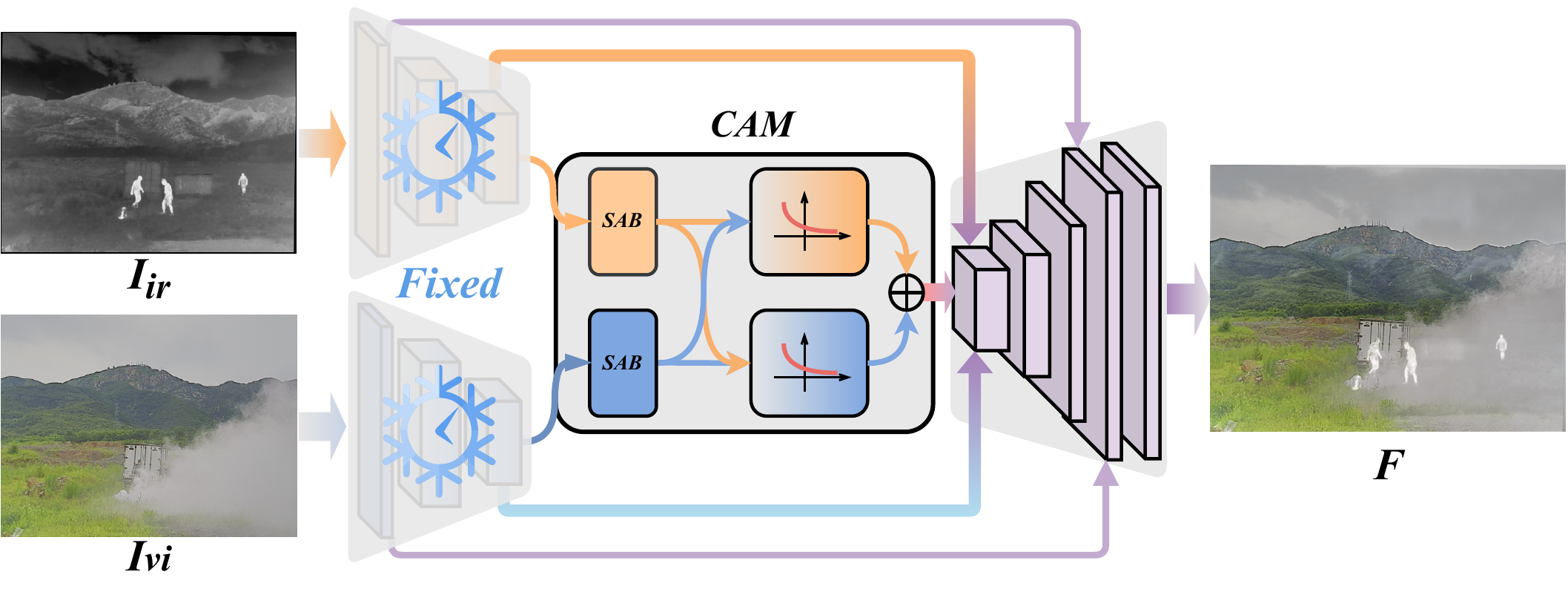}
	\caption{The second training strategy.}
	\label{fig:train_cam}
\end{figure}

In second training stage, with the fixed encoders, the proposed CAM and the decoder are trained. As shown in Fig.\ref{fig:train_cam}, between the fixed encoders and the decoder, two skip connections are also applied into the final stage.

Since the fused image should contain more complementary features and reduce redundant information from different modalities, a novel attention-based loss function is proposed to train our network. The formula of our loss function is given as follows,
\begin{eqnarray}\label{equ:loss-cam}
\begin{split}
    L_{cam} = L_{int} + w_g L_{gra}
\end{split}
\end{eqnarray}
where $w_g$ indicates the trade-off parameter between two terms, which is set to 10. $L_{int}$ and $L_{gra}$ denote the intensity loss and the gradient loss.

\textbf{The intensity loss function:} The pixel intensity indicates the main part of fused image, such as illumination, contour etc. Usually, these information do not always appear in single modality. Thus, the intensity mask is introduced into this loss function, which is given as follow,
\begin{eqnarray}\label{equ:loss-int}
\begin{split}
    L_{int} = ||F-(M_{ir}I_{ir} + M_{vi}I_{vi})||_F^2
\end{split}
\end{eqnarray}
where $F$ indicates the fused image, $M_{ir}$ and $M_{vi}$ denote the intensity masks for different modality. 

The masks are calculated as follows,
\begin{eqnarray}\label{equ:int-mask}
\begin{split}
    M_{ir} = \left\{
    \begin{aligned}
        1,& \quad loc_{ir} \geq loc_{vi}\\
        0,& \quad otherwise
    \end{aligned}
    \right.
    \quad , M_{vi} = 1 - M_{ir}
\end{split}
\end{eqnarray}
where $loc_{ir}$ and $loc_{vi}$ indicate the mean value of local patch from source images. These values are calculated as follows,
\begin{eqnarray}\label{equ:int-loc}
\begin{split}
    loc_{c} = \frac{avg_{c}}{\sum_{i\in{\{ir, vi\}}}avg_i}, \quad avg_{c} = \nabla_a I_{c}
\end{split}
\end{eqnarray}
where $avg_c$ indicates the single modality ($c\in{\{ir,vi\}}$) values obtained by mean filter $\nabla_a$, in which the kernel size is $11\times 11$.

\textbf{The gradient loss function:} Since the intensity loss function only focuses on the illumination and contour information, the gradient loss function is utilized to ensure that the detail information can be preserved. The formula of gradient loss function is given as follows,
\begin{eqnarray}\label{equ:loss-gra}
\begin{split}
    L_{gra} &= ||F - max(Clip(\nabla_g I_{ir}), Clip(\nabla_g I_{vi}))||_F^2 \\
    s.t. &\quad Clip(\cdot) = max(\cdot, 0)
\end{split}
\end{eqnarray}
where $\nabla_g$ denotes the mean filter with a small kernel size $3\times 3$. The mean filter with small kernel size can extract higher robustness features and more detail information.

\section{Experimental validation}
\label{sec-experiments}
In this section, the comparison experiments are conducted to evaluate the fusion performance of the proposed fusion method. After introducing the experimental settings, several ablation studies are performed to investigate the effect of different elements of the proposed fusion network. Several performance metrics are utilized to evaluate the fusion performance objectively.

Our network is implemented on the NVIDIA GPU (GTX 3090Ti) using PyTorch as a programming environment.

\subsection{Experimental settings}
\label{train-data}

In training phase, for the first stage (two auto-encoders), 40000 pairs of infrared and visible images are randomly chosen from the KAIST dataset. The epoch and the batch size are set to 4 and 2, respectively. For the second stage (CAM and decoder), 20000 pairs of infrared and visible images are chosen, the epoch and the batch size are set to 8 and 8, respectively. The initial learning rate is set to 0.01 and decreased by one tenth every 2 epochs. All these images are converted to gray scale and resized to $256\times 256$. 



\begin{figure}[!ht]
	\centering
	\includegraphics[width=\linewidth]{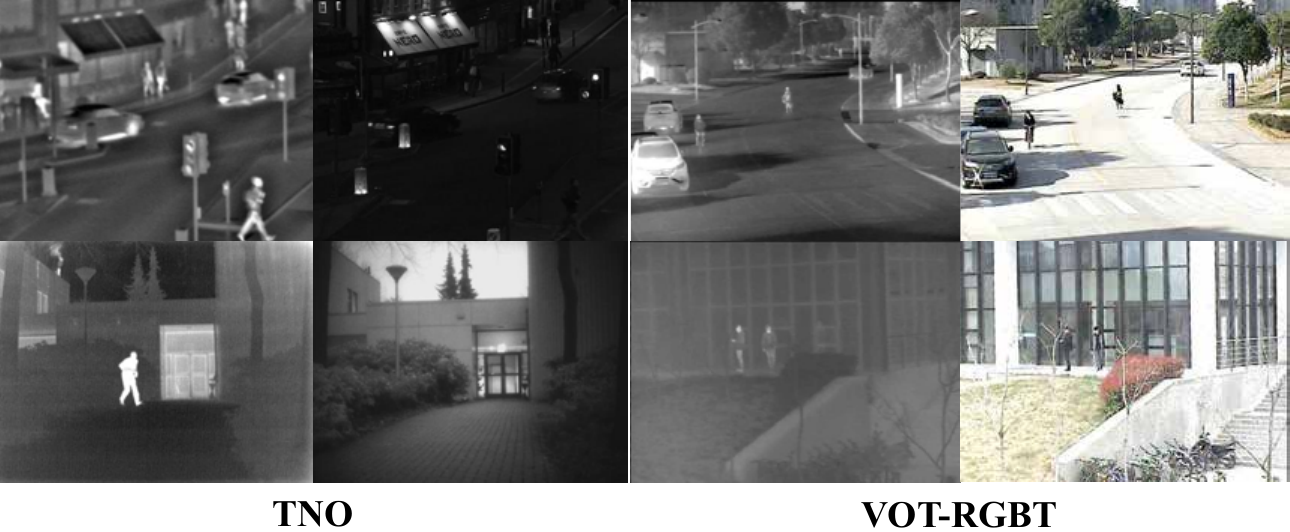}
	\caption{The examples of two datasets: TNO and VOT-RGBT.}
	\label{fig:dataset}
\end{figure}

The test images are selected from TNO \citep{tno2014} and VOT-RGBT \citep{kristan2020eighth}, comprising 21 and 40 pairs of infrared and visible images, respectively. The TNO dataset encompasses more intricate scenarios where salient objects may not always be present. In contrast, the VOT-RGBT dataset primarily concentrates on street scenarios with smaller salient targets. The examples of these two datasets are shown in Fig.\ref{fig:dataset}.

To evaluate the fusion performance of our proposed network, eight state-of-the-art fusion methods are chosen: a generative adversarial network (GAN) based fusion method (FusionGAN) \citep{ma2019fusiongan}, a unified CNN based method (IFCNN) \citep{zhang2020ifcnn}, a unified dense connection based fusion method (U2Fusion) \citep{xu2020u2fusion}, two transformer based fusion networks (YDTR \citep{tang2022ydtr}, DATFuse \citep{tang2023datfuse}), a joint down-stream tasks (saliency object detection) fusion methods (IRFS) \citep{wang2023interactively}, a semantic based fsuio methods (SemLA) \citep{xie2023semantics}, and a diffusion model based fusion network (DDFM) \citep{Zhao_2023_ICCV}. 

Furthermore, six image quality metrics are utilized to assess the objective evaluation, which includes: Entropy (En) \citep{roberts2008assessment}; Standard Deviation (SD) \citep{rao1997fibre}; Mutual Information (MI) \citep{qu2002information}; Image feature based Mutual Information ($FMI_{dct}$, $FMI_{pixel}$) \citep{haghighat2011non}; the sum of correlations of differences (SCD) \citep{aslantas2015new}.

\subsection{Ablation study}
\label{abs}
In this section, we will analyze the influence of each key part: the number of attention block, the $re\text{-}softmax$ operation, the shift operation and the CAM. Furthermore, the loss function, the fusion module and the training strategy are also analyzed.

\subsubsection{The number of ``SA'' block and ``CA'' block} 
\begin{figure}[!ht]
	\centering
	\includegraphics[width=\linewidth]{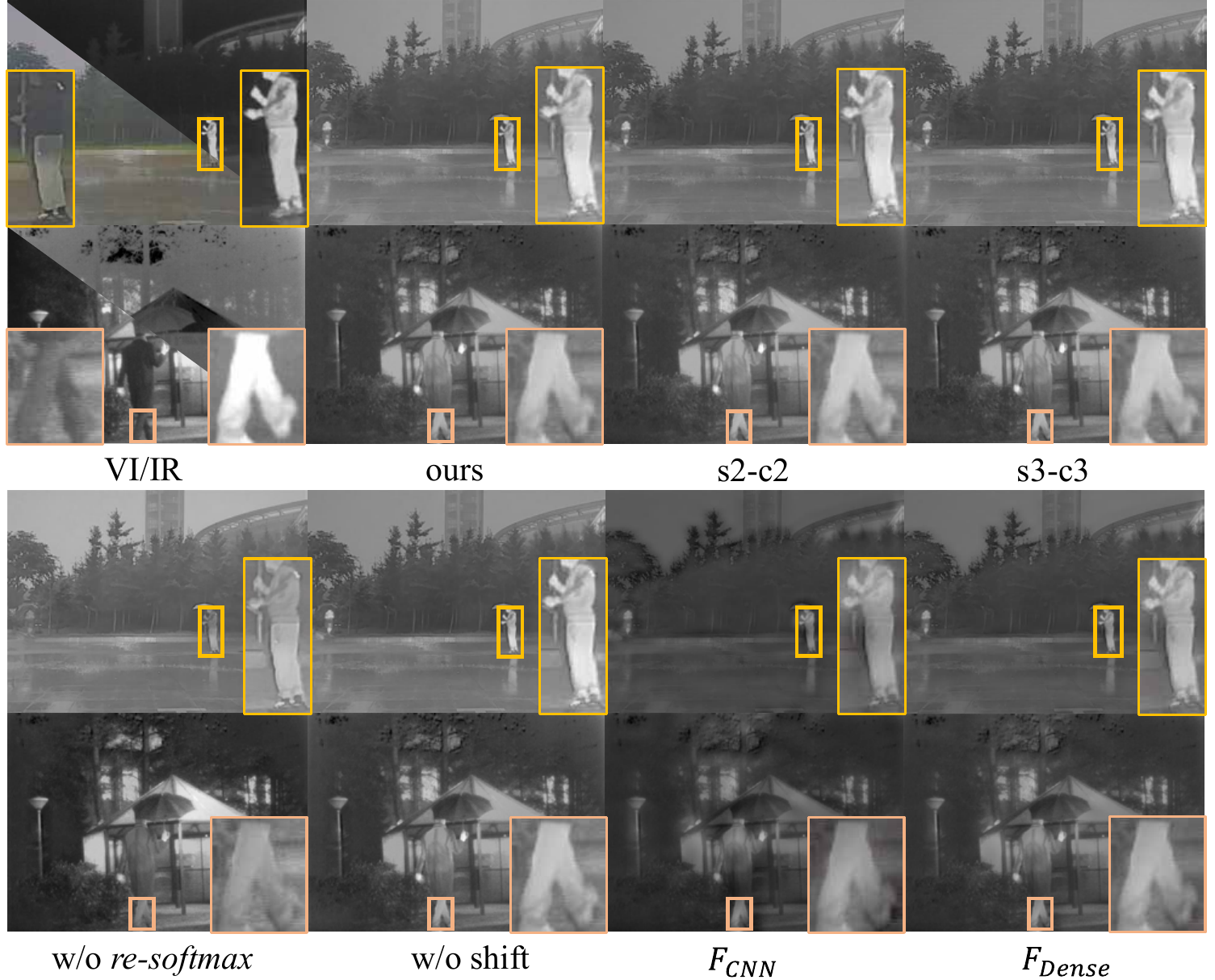}
	\caption{The results of ablation studies with different settings. The visible image is converted to gray-scale.}
	\label{fig:ablation}
\end{figure}

SA and CA indicate the self-attention module and the cross module in CAM, respectively. To find the best settings of block number, the experiments of one block (s1-c1), two blocks (s2-c2) and three blocks (s3-c3) are conducted. The visualized results and the metric values are shown in Fig.\ref{fig:ablation} and Table \ref{tab:ablation}, the best values are indicated in \textbf{bold}.

From Fig.\ref{fig:ablation}, the result obtained by one block (ours, s1-c1) contains more detail information and less artificially generated noise around the salient object (umbrella). However, the visualized performance between these results still very close. Thus, four metrics are utilized to evaluate the performance.

In Table \ref{tab:ablation}, comparing with two blocks (s2-c2) and three blocks (s3-c3), the proposed network with one SA block and one CA block (s1-c1) obtains better metric values (EN, SD, MI). Although deeper architecture has better performance in many vision tasks, it is not always correct in low-level vision task, such as image fusion. Furthermore, the proposed network is a light-weight architecture and the deep features contain less semantic features, that is why s1-c1 obtains better fusion performance.

\begin{table}[!ht]
       \tiny
       \centering
       \caption{\label{tab:ablation}The objective results of ablation studies. ``w/o cross'' means the re-softmax() operation is replaced by softmax() in CA. ``w/o shift'' indicates the shift operation is not used between two SA module. ``s1-c1'', ``s2-c2'' and ``s3-c3'' denote the number of block in SA and CA. ``$F_{CNN}$'' and ``$F_{Dense}$'' indicate that the CAM is replaced by CNN(4 conv layers) and dense architecture(1 dense block and 1 conv layer).}
       \resizebox{\linewidth}{!}{
       \begin{tabular}{c|c c c c}
         \hline
          &EN$\uparrow$       &SD$\uparrow$      &MI$\uparrow$     &$FMI_{dct}$$\uparrow$ \\
         \hline 
         ours(s1-c1)     &\textbf{6.8389}      &\textbf{73.4712}       &\textbf{13.6779}       &0.3866  \\
         \hline 
         s2-c2                &6.7282       &70.0806       &13.4563       &0.3915  \\
         s3-c3                &6.6862       &66.7879       &13.3725       &0.3868  \\
         \hline 
         w/o $re\text{-}softmax$            &6.8281       &72.5535       &13.6563       &0.3943  \\
         w/o shift            &6.7037       &68.6275       &13.4074       &\textbf{0.3974}  \\
         \hline 
         $F_{CNN}$            &6.7201       &69.6833       &13.4403       &0.3900  \\
         $F_{Dense}$          &6.7342       &70.8587       &13.4684       &0.3919  \\
         \hline 
       \end{tabular}}
\end{table}

\subsubsection{The influence of $re\text{-}softmax$ and shift operations} 

In our method, the $re\text{-}softmax(\cdot)$ operation is the key part of cross attention block, which can force network focus on the complementary (uncorrelation) information between different modalities. The shift operation is introduced to enhance the intra-features which is also applied in Swin-Transformer\citep{liu2021swin}.

As shown in Fig.\ref{fig:ablation} and Table \ref{tab:ablation}, in the absence of these two crucial operations, the fusion results exhibit a decrease in detail, and the intensity of salient objects is also diminished. Across the selected four metrics, our proposed scheme achieves three superior values (En, SD, MI), indicating that the $re\text{-}softmax(\cdot)$ and shift operations contribute to preserving more detailed information (En, SD) and enhancing complementary features (MI).

\subsubsection{The influence of CAM} 

\begin{figure}[!ht]
	\centering
	\includegraphics[width=\linewidth]{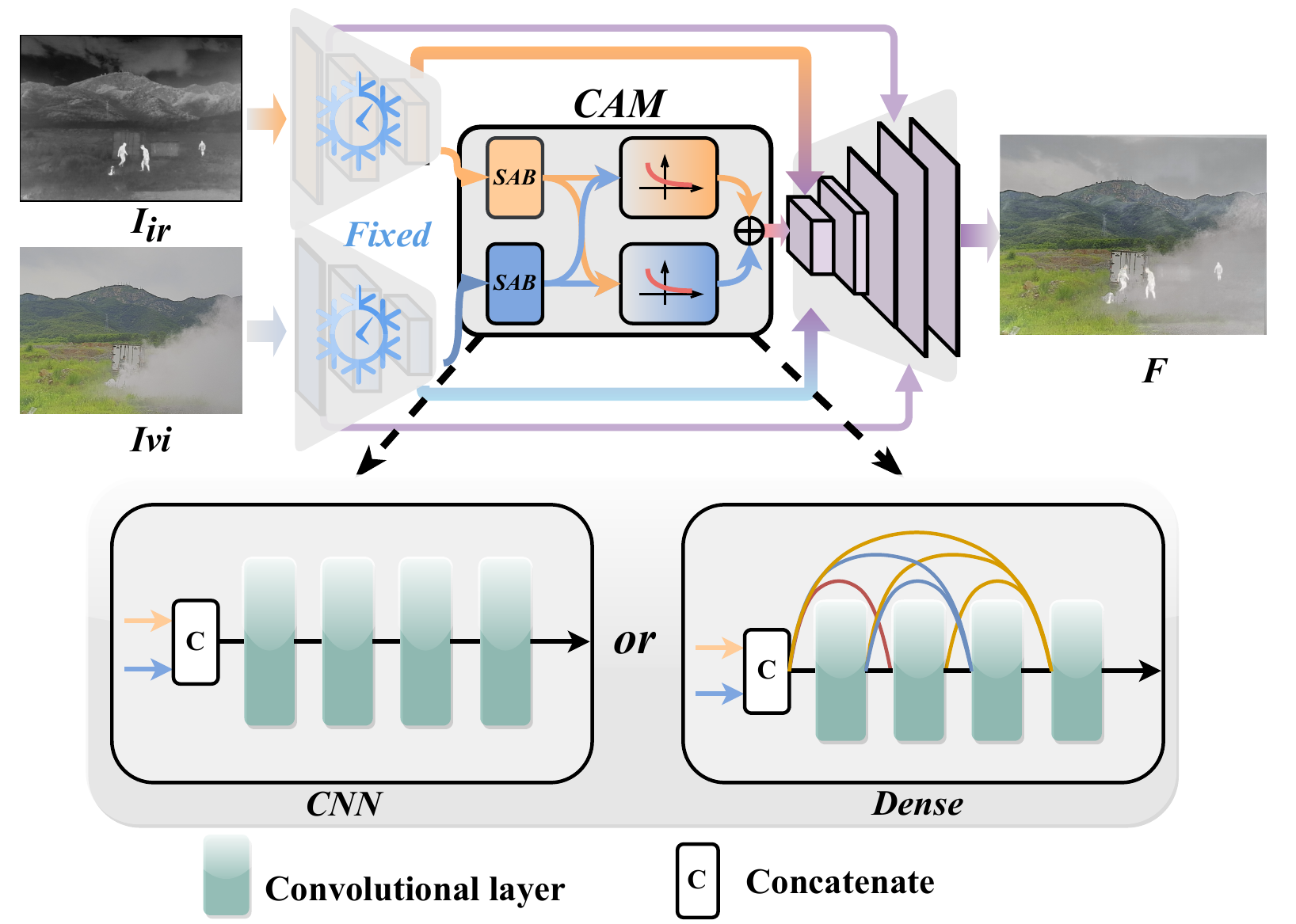}
	\caption{The proposed network with different fusion module: CAM, CNN and Dense.}
	\label{fig:fusion-module}
\end{figure}

\begin{figure}[!ht]
	\centering
	\includegraphics[width=0.9\linewidth]{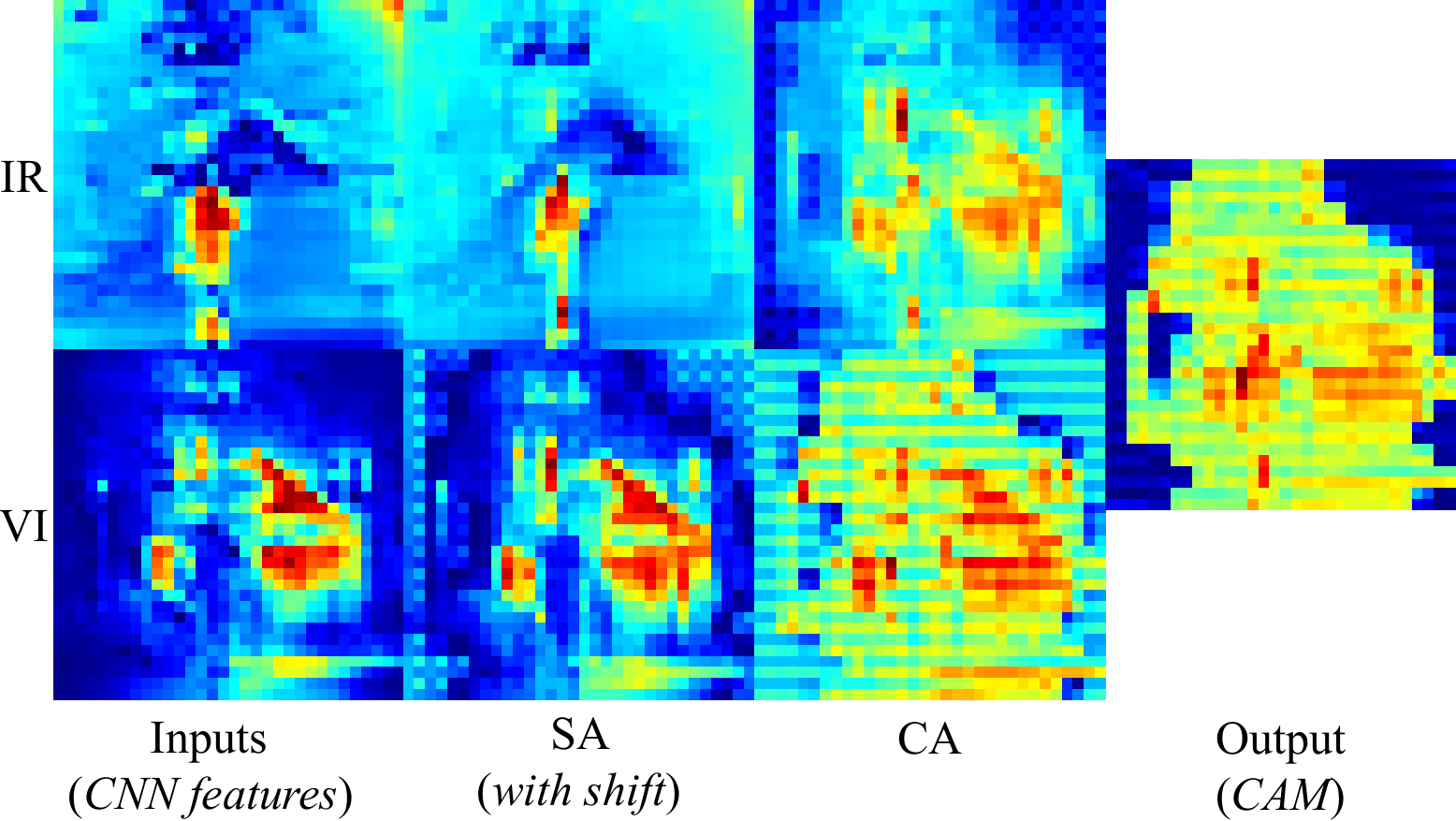}
	\caption{The visualization of middle features obtained by CNN-based encoder, SA, CA and CAM. The size of each features is $32\times 32$.}
	\label{fig:middle-feature}
\end{figure}

To evaluate the effectiveness of CAM, two architectures (CNN and Dense) are utilized to replace CAM in our fusion network. The architectures are shown in Fig.\ref{fig:fusion-module}. In Fig.\ref{fig:ablation} and Table \ref{tab:ablation}, $F_{CNN}$ and $F_{Dense}$ indicate the proposed fusion network with CNN based and dense connection based fusion module, respectively. To further analysis the influence of SA and CA in CAM, the visualization of middle features are shown in Fig.\ref{fig:middle-feature}.

As shown in Fig.\ref{fig:ablation}, comparing with $F_{CNN}$ and $F_{Dense}$, the result obtained by the proposed CAM based fusion network contains more salient features and less artifacts (background), which makes the fused image more natural. Furthermore, the objective evaluation results are shown in Table \ref{tab:ablation}, which also demonstrates that the CAM can improve the fusion performance combined with our feature extraction and image reconstruction network

As discussed in Section \ref{sec:cam}, the CAM is comprised of both the Self Attention (SA) module with shift and the Cross Attention (CA) module. The heatmaps of middle features are illustrated in Figure \ref{fig:middle-feature}. The ``Inputs'' are generated by the CNN-based encoder, highlighting the salient regions (as highlighted areas) following the source images. After the SA operations, as depicted in Figure \ref{fig:middle-feature} (SA), not only are the salient regions retained, but the finer details (background) are also enriched within each modality. Thanks to the CA operation, the complementary regions in each branch (IR, salient parts, and VI, background) are amplified through the ``$re\text{-}softmax(\cdot)$''. The final addition operation culminates in the acquisition of the enhanced features via our CAM. The visual results obtained by different architectures are also shown in Fig.\ref{fig:ablation} (ours, $F_{CNN}$ and $F_{Dense}$).

These observations indicate that our proposed CAM effectively augments the complementary features within multimodal images, ensuring the preservation of both salient objects and detailed information.

\subsubsection{Analysis for loss function and fusion module} 

\begin{figure}[!ht]
	\centering
	\includegraphics[width=\linewidth]{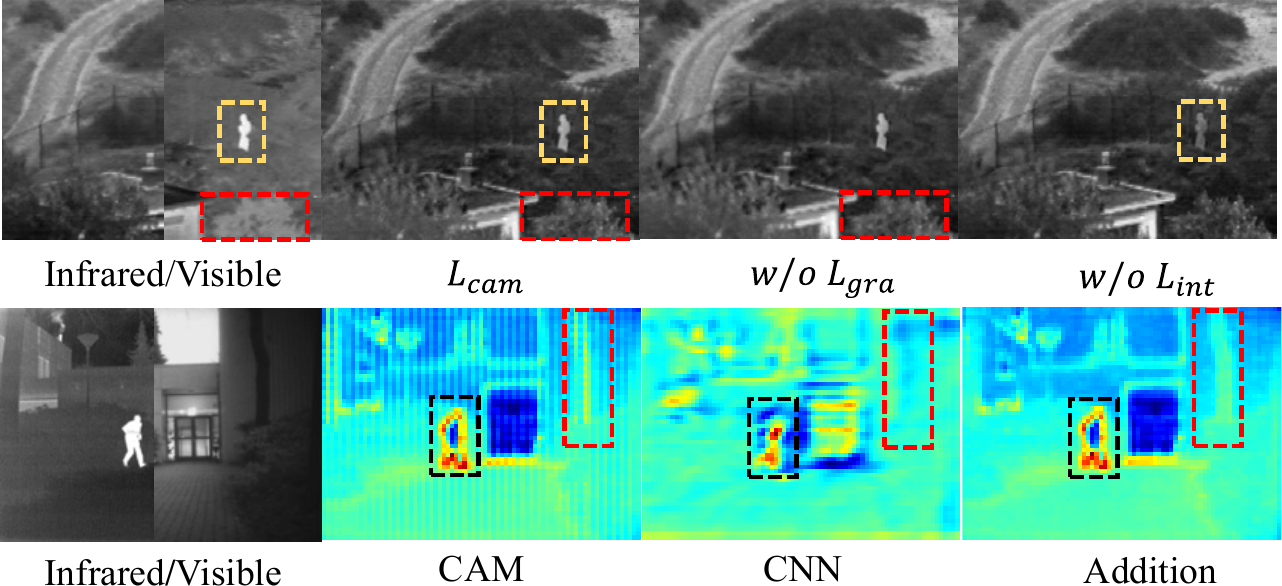}
	\caption{Fusion results obtained by different loss function settings and the visualization of fused features obtained by different fusion strategies.}
	\label{fig:loss-feature}
\end{figure}

In second training stage, the loss function ($L_{cam}$) contains two items: the pixel intensity part ($L_{int}$) and the gradient part ($L_{gra}$). In Fig.\ref{fig:loss-feature} (first row), ``w/o $L_{gra}$'' indicates only the pixel intensity part is utilized to train our network and ``w/o $L_{int}$'' means only the gradient part is used. 

As shown in Fig.\ref{fig:loss-feature} (first row), in comparison with ``$L_{cam}$'', the fusion result obtained by ``w/o $L_{gra}$'' exhibits a reduction in detailed information, as indicated by the red box. For ``w/o $L_{int}$'', the intensity of the salient object in the fused image decreases, as highlighted in the yellow box, which is deemed unacceptable for the image fusion task. These observations underscore the effectiveness of our proposed loss function in preserving both detailed information and salient pixel intensity.

To analysis the performance of fusion module, the visual results of fused deep features are shown in Fig.\ref{fig:loss-feature} (second row). These heat maps are calculated by average feature maps across channel dimensions. To evaluate the fusion performance, two classical fusion modules are chosen: (1) a light-weight CNN based fusion module (CNN), and (2) additional operation based fusion module (Addition).

In Fig.\ref{fig:loss-feature} (second row), the feature map derived from the CNN exhibits a higher presence of redundant features. In contrast to the CNN module, the CAM proves adept at preserving more structural information from source inputs, enhancing salient objects (as highlighted in the black and red boxes). Moreover, the CAM effectively amplifies complementary regions from multi-modality inputs (as indicated by the red box), outperforming the Addition method. These observations affirm that our proposed fusion module (CAM) excels in augmenting complementary features and structural information while mitigating the presence of redundant features.

\subsubsection{Analysis for training strategies} 

\begin{figure}[!ht]
	\centering
	\includegraphics[width=\linewidth]{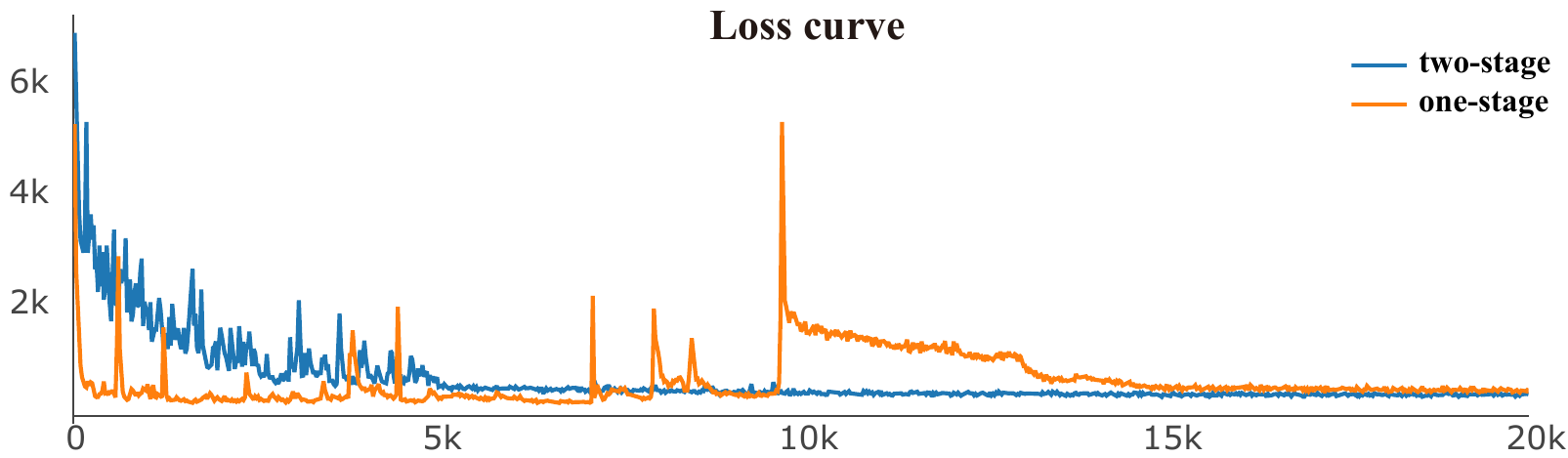}
	\caption{Fusion results with different loss function settings and the visualization of fused features with different fusion strategies (``two-stage'' and ``one-stage'').}
	\label{fig:loss-curve}
\end{figure}

In this section, we will analyze the impact of different training strategies, namely the ``two-stage'' and ``one-stage'' strategies. ``two-stage'' indicates the training strategy utilized in our proposed fusion framework, ``one-stage'' means two encoders, CAM and decoder are trained together with the proposed loss function. The loss curve and metrics values (on TNO) are shown in Fig.\ref{fig:loss-curve} and Table \ref{tab:ablation-stage}.

\begin{table}[!ht]
       \tiny
       \centering
       \caption{\label{tab:ablation-stage} The objective results of different training strategies (``two-stage'' and ``one-stage'') on TNO.}
       \resizebox{\linewidth}{!}{
       \begin{tabular}{c|c c c c}
         \hline
          &EN$\uparrow$       &SD$\uparrow$      &MI$\uparrow$     &$FMI_{dct}$$\uparrow$ \\
         \hline 
         two-stage    &\textbf{6.8389}      &\textbf{73.4712}       &\textbf{13.6779}       &0.3866  \\
         one-stage &6.6743       &68.5877      &13.3486      &\textbf{0.3900} \\
         \hline
       \end{tabular}}
\end{table}

As shown in Fig.\ref{fig:loss-curve}, with the utilization of the proposed loss function ($L_{cam}$), both of these training strategies converge to a stable value. Nevertheless, under identical settings, the ``two-stage'' approach exhibits faster convergence and a smaller loss value compared to the ``one-stage'' strategy.

Moreover, to ascertain the superior training strategy, we conducted comparative experiments on TNO. Table \ref{tab:ablation-stage} reveals that the ``two-stage'' approach attains the three highest values, signifying that a fusion model based on the two-stage training strategy yields superior fusion performance. Based on the efficiency of training strategy and the fusion performance, in our proposed framework, we choose two-stage training strategy to train our network.

\begin{figure*}[!ht]
	\centering
	\includegraphics[width=\linewidth]{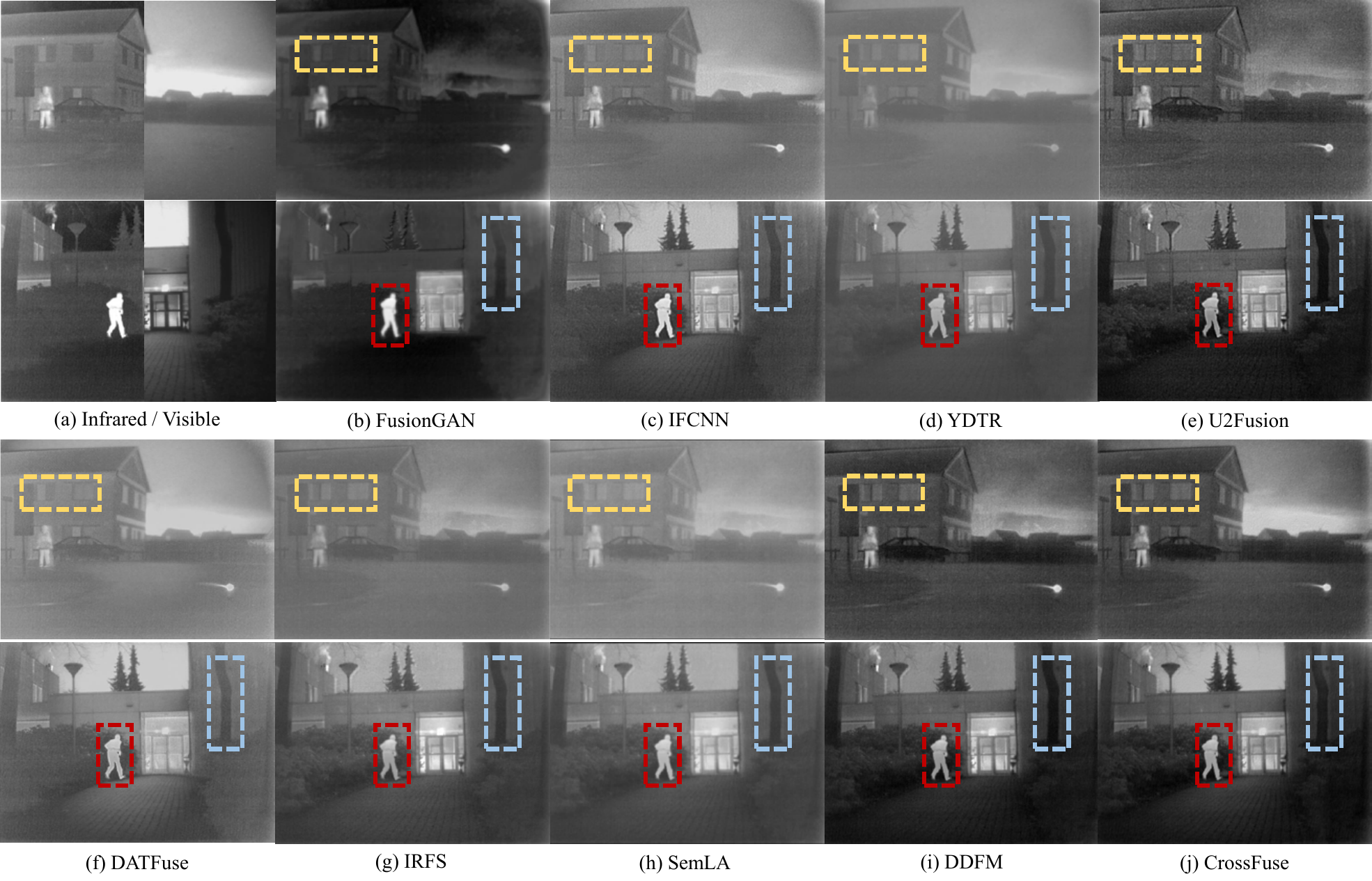}
	\caption{The fusion results obtained by compared fusion methods and the proposed method on TNO (``yard'' and ``man''). (a) Infrared and Visible images; (b) FusionGAN; (c) IFCNN; (d) YDTR; (e) U2Fusion; (f) DATFuse; (g) IRFS; (h) SemLA; (i) DDFM; (i) CrossFuse (\emph{the proposed}).}
	\label{fig:results-tno}
\end{figure*}

\begin{figure*}[!ht]
	\centering
	\includegraphics[width=\linewidth]{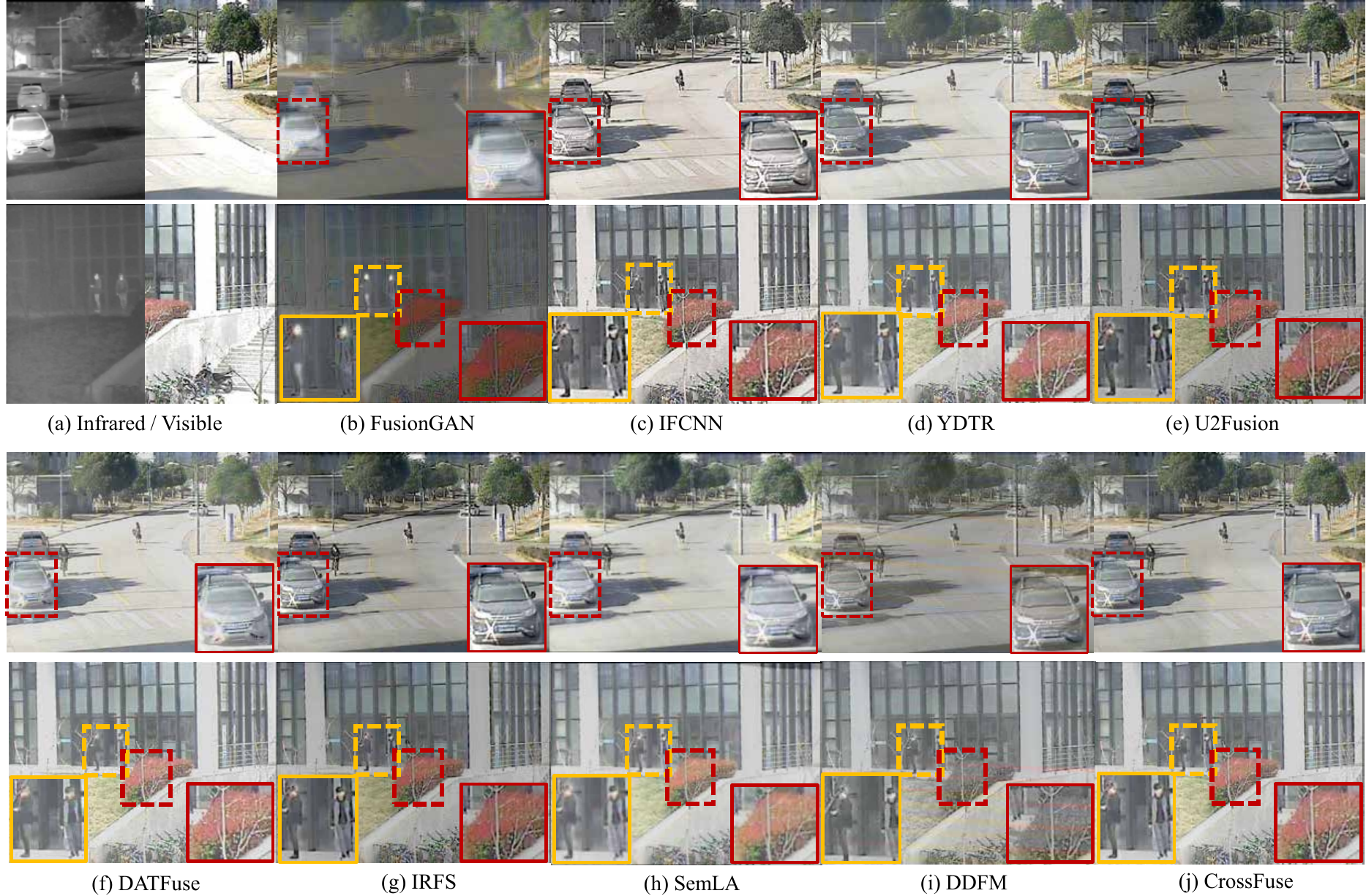}
	\caption{The fusion results obtained by compared fusion methods and the proposed method on VOT-RGBT (``crossroad'' and ``two-man''). (a) Infrared and Visible images; (b) FusionGAN; (c) IFCNN; (d) YDTR; (e) U2Fusion; (f) DATFuse; (g) IRFS; (h) SemLA; (i) DDFM; (i) CrossFuse (\emph{the proposed}).}
	\label{fig:results-vot2}
\end{figure*}

\begin{table*}[!ht]
       \tiny
       \centering
       \caption{\label{tab:result-tno}The average metrics values obtained by the existing fusion methods and the proposed network on TNO.}
       \resizebox{0.85\linewidth}{!}{
       \begin{tabular}{c|c| c c c c c c}
         \hline
        &year  &EN$\uparrow$           &SD$\uparrow$          &MI$\uparrow$           &$FMI_{dct}$$\uparrow$      &$FMI_{pixel}$$\uparrow$    &SCD$\uparrow$ \\
        \hline
        FusionGAN \citep{ma2019fusiongan}     &2020  &6.3629      &54.3575     &12.7257     &0.3634      &0.8906      &1.4569 \\
        IFCNN \citep{zhang2020ifcnn}         &2021  &6.5955      &\emph{\color{red}{\underline{66.8758}}}     &13.1909     &\emph{\color{red}{\underline{0.3738}}}      &0.9009      &1.7138 \\
        YDTR \citep{tang2022ydtr}          &2022  &6.2268      &51.4882     &12.4536     &0.3483      &0.8992      &1.6106 \\
        U2Fusion \citep{xu2020u2fusion}      &2022  &\emph{\color{red}{\underline{6.7571}}}      &64.9116     &\emph{\color{red}{\underline{13.5142}}}     &0.3406      &0.8936      &\textbf{1.7984} \\
        DATFuse \citep{tang2023datfuse}       &2023  &6.3206      &56.0363     &12.6412     &0.2738      &0.8807      &1.5240 \\
        IRFS \citep{wang2023interactively}     &2023 &6.43326       &59.13428      &12.86652      &0.37997       &0.90520       &1.74604 \\
        SemLA \citep{xie2023semantics}        &2023 &6.52166       &63.92465      &13.04331      &0.16495       &0.90865       &1.53009 \\
        DDFM \citep{Zhao_2023_ICCV}       &2023 &6.72427       &66.64661      &13.44855      &0.21777       &0.88257       &1.54674 \\
        CrossFuse     &\emph{ours}      &\textbf{6.8389}      &\textbf{73.4712}     &\textbf{13.6779}     &\textbf{0.3866}      &\emph{\color{red}{\underline{0.9044}}}      &\emph{\color{red}{\underline{1.7659}}} \\
         \hline 
       \end{tabular}}
\end{table*}

\subsection{Fusion results analysis}
\label{abs}
In this section, five state-of-the-art fusion methods and six metrics are chosen to evaluate the fusion performance of our proposed fusion network. The comparison experiments are conducted on two public fusion datasets (TNO \citep{tno2014} and VOT-RGBT \citep{kristan2020eighth}). The examples ``yard'' and ``man'' from TNO, the examples ``outdoors'' and ``two-man'' from VOT-RGBT, are chosen to show the visual results.

\subsubsection{Fusion Results on TNO}
To evaluate the fusion performance on TNO \citep{tno2014}, 21 pairs of infrared and visible images are selected. The fusion results obtained by the proposed method and other existing fusion methods on TNO (``yard'' and ``man'') are shown in Fig.\ref{fig:results-tno}.

Comparing with the GAN-based method (FusionGAN \citep{ma2019fusiongan}), CNN-based method (IFCNN \citep{zhang2020ifcnn}) and dense connection-based method (U2Fuison \citep{xu2020u2fusion}), the fused image obtained by our proposed method contains more detail information (Fig.\ref{fig:results-tno}, yellow box). Moreover, our proposed method can generate clearer fused image than the transformer-based method (YDTR \citep{tang2022ydtr} and DATFuse \citep{tang2023datfuse}) and two down-stream task based methods (IRFS \citep{wang2023interactively} and SemLA \citep{xie2023semantics}). For the diffusion model based fusion method (DDFM \citep{Zhao_2023_ICCV}), the proposed method obtains the comparable visual results on TNO.

To assess the fused image quality objectively, six metrics are selected. The metrics values are shown in Table \ref{tab:result-tno}, the best values are denoted in \textbf{blob} and the second-best values are denoted in \emph{\color{red}{\underline{italic and red}}}. In Table \ref{tab:result-tno}, compared with other state-of-the-art fusion methods, our proposed method (CrossFuse) achieves four best values (EN, SD, MI, and $FMI_{dct}$) and two second-best value ($FMI_{pixel}$ and SCD), which means the CrossFuse can preserve more complementary information from pixel-level (EN and SD) and feature-level (MI and $FMI_{dct}$). The above observations indicate that the CAM can maintain more complementary information from source images.

\begin{table*}[!ht]
       \tiny
       \centering
       \caption{\label{tab:result-votrgbt}The average metrics values obtained by the existing fusion methods and the proposed network on VOT-RGBT.}
       \resizebox{0.85\linewidth}{!}{
       \begin{tabular}{c|c| c c c c c c}
          \hline 
           &year  &EN$\uparrow$           &SD$\uparrow$          &MI$\uparrow$           &$FMI_{dct}$$\uparrow$      &$FMI_{pixel}$$\uparrow$    &SCD$\uparrow$ \\
          \hline 
          FusionGAN \citep{ma2019fusiongan}    &2020    &6.5203        &62.8494       &13.0406       &0.3646        &0.8912        &1.3748 \\ 
        IFCNN \citep{zhang2020ifcnn}           &2021    &6.7411        &76.2492       &13.4821       &0.3736        &0.9047        &1.6686 \\ 
        YDTR \citep{tang2022ydtr}              &2022    &6.4012        &62.4483       &12.8024       &0.3461        &0.9051        &1.5624 \\ 
        U2Fusion \citep{xu2020u2fusion}        &2022    &\textbf{6.9487}        &\emph{\color{red}{\underline{76.7838}}}       &\textbf{13.8973}       &0.3364        &0.8970        &\textbf{1.7479} \\ 
        DATFuse \citep{tang2023datfuse}        &2023    &6.4580        &62.3694       &12.9160       &0.2745        &0.8843        &1.4845 \\ 
        IRFS \citep{wang2023interactively}     &2023    &6.6071        &67.5912       &13.2141       &\emph{\color{red}{\underline{0.3740}}}        &0.9059        &\emph{\color{red}{\underline{1.7117}}} \\ 
        SemLA \citep{xie2023semantics}         &2023    &6.6757        &71.5133       &13.3513       &0.1618        &\textbf{0.9074}        &1.5475 \\ 
        DDFM \citep{Zhao_2023_ICCV}            &2023    &6.8214        &73.2632       &13.6428       &0.1805        &0.8772        &1.5112 \\ 
        CrossFuse                 &\emph{ours}    &\emph{\color{red}{\underline{6.8908}}}        &\textbf{77.1780}       &\emph{\color{red}{\underline{13.7816}}}       &\textbf{0.3827}        &\emph{\color{red}{\underline{0.9061}}}        &1.6635 \\ 
          \hline 
       \end{tabular}}
\end{table*}

\subsubsection{Fusion results on VOT-RGBT}
In VOT-RGBT, 40 pairs of infrared and visible images are selected from VOT-RGBT \citep{kristan2020eighth} and TNO \citep{tno2014}. Since the visible image is in RGB space, it is converted into YCrCb color space. ``Y'' indicates the luminance, ``Cr'' and ``Cb'' denote the chrominance. To obtain the RGB fused image, ``Y'' and the infrared image (gray-scale) are fused by the fusion method firstly. Then, ``Cr'' and ``Cb'' are directly combined with the fused gray-scale image to generate the final fused image in YCrCb space. Finally, the fused image is converted to RGB space.

In Fig.\ref{fig:results-vot2}, two pairs of infrared and visible images, ``crossroad''  and ``two-man'', are chosen to demonstrate the visual results generated by the existing fusion methods and the proposed method. Comparing with IFCNN \citep{zhang2020ifcnn}, YDTR \citep{tang2022ydtr}, U2Fusion \citep{xu2020u2fusion} and DATFuse \citep{tang2023datfuse}, the infrared objects are enhanced by the proposed CrossFuse (Fig.\ref{fig:results-vot2}, red box and yellow box). This observation indicates that the novel cross attention mechanism can enhance more complementary features between different modalities compared with state-of-the-art fusion methods. Furthermore, comparing with FusionGAN \citep{ma2019fusiongan}, IRFS \citep{wang2023interactively}, SemLA \citep{xie2023semantics} and DDFM \citep{Zhao_2023_ICCV}, our proposed network preserves more detail information (Fig.\ref{fig:results-vot2} ``two-man'', yellow box) while obtaining the comparable salient objects.

The average values\footnote{These metrics are calculated under the gray-scale space, which means the visible images are converted to gray-scale firstly.} of six metrics are shown in Table \ref{tab:result-votrgbt}, the best values and the second-best values are denoted in \textbf{blob} and \emph{\color{red}{\underline{italic and red}}}, respectively. The proposed fusion method, CrossFUse, obtains two best values (SD and $FMI_{dct}$) and three second-best values (EN, MI, $FMI_{pixel}$). Although the proposed method dose not achieve all best values, comparing with the state-of-the-art fusion methods, it still achieves comparable metrics values and even better fusion performance in image sharpness (SD). These observations indicate that our proposed method obtains better fusion performance in both visual evaluation and objective evaluation.

\section{Conclusions}
\label{sec-con}

The uncorrelation (complementary) is the key to multimodal image fusion task, which needs to be paid more attention. Unfortunately, the existing transformer based methods ignore this limitation. Thus, a novel hybrid (CNN and transformer) fusion network (CrossFuse) is introduced, in which a new cross attention mechanism (CAM) is proposed and applied to fusion module. The key part of CAM is $re\text{-}softmax(\cdot)$ operation which is utilized to enhance the complementary features between different modalities and reduce the redundant information. Moreover, a simple yet efficient loss function is also proposed to force the fused image contains more salient features and detail information from source images. The experimental results on public datasets show that our proposed fusion network demonstrates better fusion performance than the start-of-the-art fusion methods.

While the proposed cross-attention mechanism proves simple yet efficient in the image fusion task, it has limitations in significantly enhancing fusion performance within the transformer framework. A possible research direction involves incorporating additional machine learning methods, such as sparse representation and metric learning, to augment the effectiveness of the cross-attention mechanism. Future efforts will be directed towards exploring and implementing these solutions.

\section*{Acknowledgement}

This work was supported by the National Natural Science Foundation of China (62202205, 62332008, 62020106012), National Key Research and Development Program of China (2023YFE0116300), and the Fundamental Research Funds for the Central Universities (JUSRP123030).
\bibliographystyle{elsarticle-num-names} 

\bibliography{crossfuse}

\end{document}